\PassOptionsToPackage{dvipsnames,table,xcdraw}{xcolor}
\documentclass[10pt, a4paper]{article}
\pdfoutput=1
\usepackage{lrec2022}
\usepackage{fancybox, graphicx}
\usepackage{tabularx}
\usepackage{soul}
\usepackage{tikz}
\usetikzlibrary{shapes,shadows,arrows,positioning,fit,calc,matrix}
\usepackage{tabu}
\usepackage{pifont}
\usepackage{multirow}
\usepackage{pgfplots}
\usepackage{caption}
\usepackage{pgfplotstable}
\pgfplotsset{compat=1.8}
\usepgfplotslibrary{statistics}
\usepackage{paralist}
\usepackage{booktabs}
\usepackage{tabularx}
\usepackage{multirow}
\usepackage{tikz-dependency}
\usepackage{titlesec}
\titleformat{\section}{\normalfont\large\bfseries\center}{\thesection.}{1em}{}
\titleformat{\subsection}{\normalfont\SmallTitleFont\bfseries\raggedright}{\thesubsection.}{1em}{}
\titleformat{\subsubsection}{\normalfont\normalsize\bfseries\raggedright}{\thesubsubsection.}{1em}{}
\renewcommand\thesection{\arabic{section}}
\renewcommand\thesubsection{\thesection.\arabic{subsection}}
\renewcommand\thesubsubsection{\thesubsection.\arabic{subsubsection}}

\usepackage[utf8]{inputenc}
\usepackage{hyperref}
\usepackage{xstring}
\usepackage{color}

\usepackage{enumitem,amssymb}
\newlist{todolist}{itemize}{2}
\setlist[todolist]{label=$\square$}

\newcommand{\surround}[1]{$\big[$#1$\big]$}
\newcommand{\stack}[2]{$\stackunder{\text{#1}}{\text{\tiny\sf #2}}$}
\newcommand{\medCond}[1]{\surround{\textcolor{orange}{\stack{#1}{\textsc{medCond}}}}}
\newcommand{\treatment}[1]{\surround{\textcolor{Blue}{\stack{#1}{\textsc{treatment}}}}}
\newcommand{\sympt}[1]{\surround{\textcolor{ForestGreen}{\stack{#1}{\textsc{sympt/side-effect}}}}}
\newcommand{\other}[1]{\surround{\textcolor{black}{\stack{#1}{\textsc{other}}}}}
\usepackage{amsmath}
\usepackage{stackengine}

\usepackage{xspace}
\newcommand{\F}{F$_1$\xspace}

\title{CoVERT: A Corpus of Fact-checked Biomedical COVID-19 Tweets\\}
\name{Isabelle Mohr, Amelie Wührl, Roman Klinger} 

\address{Institut für Maschinelle Sprachverarbeitung, University of Stuttgart, Germany \\
  \{isabelle.mohr, amelie.wuehrl, roman.klinger\}@ims.uni-stuttgart.de\\}

\abstract{ Over the course of the COVID-19
  pandemic, large volumes of biomedical information concerning this
  new disease have been published on social media. Some of this
  information can pose a real danger to people's health, particularly
  when false information is shared, for instance recommendations on how
  to treat diseases without professional medical advice. Therefore,
  automatic fact-checking resources and systems developed specifically
  for the medical domain are crucial. While existing fact-checking
  resources cover
  COVID-19-related information in news or quantify the amount of
  misinformation in tweets, there is no dataset providing fact-checked
  COVID-19-related Twitter posts with detailed annotations for
  biomedical entities, relations and relevant evidence. We contribute
  CoVERT, a fact-checked corpus of tweets with a focus on
  the domain of biomedicine and COVID-19-related (mis)information. The
  corpus consists of 300 tweets, each annotated with medical named entities
  and relations. We employ a novel crowdsourcing methodology to
  annotate all tweets with fact-checking labels and supporting
  evidence, which crowdworkers search for online. This
  methodology results in moderate inter-annotator
  agreement. Furthermore, we use the retrieved evidence extracts as
  part of a fact-checking pipeline, finding that the real-world
  evidence is more useful than the knowledge indirectly available in
  pretrained language models.%
 \\ \newline  }

\begin{document}

\maketitleabstract

\section{Introduction}

The COVID-19 pandemic has elicited a global response in the scientific
community, leading to a burst of new information regarding the
pathophysiology of the virus, new treatments for infected patients as
well as vaccines in production \cite{chahrour-2020-bibliometric}.  At
the same time, large volumes of information about COVID-19 have also
been published on social media platforms
\cite{yang-2020-lowcredInfoTwitter,kouzy-2020-coronagoesviral,singh2020look,shahi2020exploratory},
which are known to reach large networks of individuals in short
amounts of time \cite{kouzy-2020-coronagoesviral}. This reach is
paramount to curbing the spread of a virus like SARS-CoV-2. Although
much of the information circulating online is potentially useful, some
of it, in particular misinformation and disinformation, can also pose
a great danger \cite{soltaninejad-2020-methanol}. Incorrect health
advice can not only put individuals at risk, false and misleading
information may also be detrimental to efforts in controlling the
pandemic.  Therefore, the stream of information on social media sites
requires critical thought and fact-checking.

Manual fact-checking, however, is time-consuming and expensive. One
way to perform the task automatically is to use machine learning to
classify a claim to be true or false. An
intermediate step to exploit potential evidence is to automatically decide for
claim-evidence pairs if the evidence supports or refutes
the claim (or whether that information is insufficient to rule a verdict)
\cite{vlachos-riedel-2014-fact}. The growing need for fact-checking of
COVID-19-related information has been met with releases of corpora of
fact-checked COVID-19 claims
\cite{shahi-2020-fakecovid,shahi2020exploratory}, and databases
containing known facts concerning COVID-19 and related information
\cite{reese-2021-kgCovid,fernandez-2020-covid19kg}. However, to the
best of our knowledge, there is no resource that specifically
addresses the truthfulness of biomedical information relating to
COVID-19 circulating on Twitter.

\begin{table}
  \centering\small
  \setlength{\tabcolsep}{4pt}
  \renewcommand{\arraystretch}{0.7}
  \renewcommand\tabularxcolumn[1]{m{#1}}
  \begin{tabularx}{\columnwidth}{lp{0.35\textwidth}}\toprule
	  Variable & Values \\\cmidrule(lr){1-2}
    NER Tweet & \begin{dependency}[edge style={Gray}, label style={gray!60, fill=gray!60,font=\sf,text=black}]
      \begin{deptext}
        \other{5G networks} \& caused \& \medCond{covid} \& .\\
      \end{deptext}
      \depedge[edge unit distance=0.5ex]{1}{3}{cause\_of}
    \end{dependency} \\\cmidrule(lr){1-2}
    Verdict & \textsc{refutes} \\\cmidrule(lr){1-2}
    URL & \url{https://www.muhealth.org/} $\ldots$ \\\cmidrule(lr){1-2}
    Evidence & There are two types of conspiracy associated with 5G-COVID-19. One version suggests that radiation from 5G lowers your immune system, which makes you more susceptible to the virus (Shultz, 2020). The idea that $\ldots$\\\bottomrule
  \end{tabularx}
  \caption{Example instance from the CoVERT corpus, with
    NER and crowdsourced fact-checking annotations,
    including textual evidence and the source URL.}
  \label{tab:manual_scispacy_NER}
\end{table}

We fill this gap by creating a corpus of fact-checked biomedical
tweets, the Covid VERified Tweet (CoVERT) corpus\footnote{The CoVERT
  corpus and supplementary material is available at
  \url{https://www.ims.uni-stuttgart.de/data/bioclaim}.},
to facilitate the fact-checking task in this domain. The dataset
consists of 300 tweets containing real-world claims about COVID-19. We
annotate the tweets with two types of information: First, we annotate
each tweet with the medical entities \textit{Medical condition},
\textit{Treatment}, \textit{Symptom/Side-effect}, \textit{Other} and
annotate how they relate to each other (\textit{cause\_of},
\textit{causative\_agent\_of}, \textit{not\_cause\_of}). Secondly,
crowdworkers verify the medical claim within the tweet. To the best of
our knowledge, employing crowdworkers to perform biomedical
fact-checking on Twitter posts has not been done before, though
crowdworkers have completed subtasks for veryfing COVID-19-related
Reddit posts.
Three annotators research a given claim and
provide substantiating evidence from the web along with their verdict
(\textsc{supports}, \textsc{refutes} and \textsc{not enough
information (nei)}). Table~\ref{tab:manual_scispacy_NER} shows an
example instance from our dataset.

In an exploratory analysis of this dataset we find our novel
methodology of crowdsourced fact-checking to be effective for this
task, with moderate agreement between annotators on their
verdicts.
Finally, we explore to which extent textual evidence extracts provided
by the annotators help inform fact-checking systems when making a
prediction. We find that real evidence provides our pipeline with more
useful information than what is available implicitly in the pretrained
language model BERT \cite{devlin-etal-2019-bert}.
However, the \textsc{not enough information} class
inherently has no substantiating evidence, making this class a
challenge for this approach.

\section{Related Work}
In recent years, automated fact-checking has increasingly come into
focus \cite{thorne-vlachos-2018-automated}. Systems can be grouped
into approaches with and without access to external evidence. The
former need to combine information from other texts or
structured resources with the claim, while the latter
rely on linguistic patterns that signal false information.

\subsection{Fact-Checking without External Evidence}
\label{rel-work_without-evidence}
Previous research that automatically check claims without evidence
make use of, i.a. emotion patterns \cite{Giahanou2019LeveragingES} or
surface-level linguistic properties \cite{rashkin}. These can be
categorized as stylometric approaches \cite{schuster-stylography}.
Based on the finding that language models capture relational
information contained within the data they are trained on,
\newcite{lee2020language} include language models in the fact-checking
pipeline itself. Tapping into the implicit knowledge aggregated over
very large pretraining datasets, they use BERT
\cite{devlin-etal-2019-bert} to create an evidence text by unmasking
an entity in the original claim. This is passed to an entailment model
that predicts whether the language model's evidence supports or
refutes the claim.

Both the stylometric and the language model-based approaches are
limiting. The first group fails to recognize well-presented false
information that do not share the surface level characteristics of
false news. When using a language model to extract evidence, the
models also likely propagate biases learned from the training data
\cite{vlachos-2021-survey}. Additionally, such models are biased
towards particular prompts which are used to query and extract the
implicit knowledge from the model \cite{sung2021language}.

\subsection{Evidence-based Approaches}
Both lines of research described in Section
\ref{rel-work_without-evidence} are in contrast with the way
journalists approach fact-checking, who rather seek substantiating
external evidence. To go beyond the claim itself when making a
verdict prediction, \newcite{vlachos-riedel-2015-identification} and
\newcite{ciampaglia-2015-comp} use the structured nature of
information available in knowledge bases, while others retrieve
evidence extracts from resources like news articles and websites
online \cite{popat2018,hanselowski-etal-2019-richly}. This allows to
fact-check claims with probabilistic estimates
\cite{vlachos-riedel-2015-identification} or to use a knowledge graph
as a topology to predict how likely a claim is to be true
\cite{ciampaglia-2015-comp}.

In an effort to mirror the traditional process of journalistic
fact-checking, \newcite{popat2018} retrieve substantiating articles
that provide evidence for or against a claim. Further, they
incorporate a credibility assessment of the claim's source in their
model.

In the related task of stance detection,
\newcite{ferreira-vlachos-2016-emergent} use news headlines as
evidence for claims. Other approaches extract the summaries of
articles rather than their headlines as evidence for claims
\cite{hanselowski-etal-2019-richly,alhindi-etal-2018-evidence},
filtering out sentences that are irrelevant in order to create more
fine-grained substantiating evidence. Similarly,
\newcite{wadden2021longchecker} make use of the evidence contained
within the abstracts of research papers in order to fact-check
scientific claims.

Retrieving evidence allows systems to make informed decisions based on
external sources which helps overcome the limitations claim-focused
models face. In addition, evidence can also provide the end-user with
an explanation for the fact-checking verdict \cite{kotonya-toni_2020}.

\subsection{Fact-Checking Corpora}

Recently, the task of fact-checking has received much attention
\cite{thorne-vlachos-2018-automated}, resulting in an influx of
published datasets. The Fact Extraction and VERification (FEVER)
\cite{Thorne18FEVER} dataset is one of the largest available corpora,
consisting of 185,445 claims generated from Wikipedia. Claims are
manually annotated (labels: \textsc{supported}, \textsc{refuted},
\textsc{not enough info}) and accompanied by an evidence sentence from
the same original Wikipedia article. In the political domain,
previously fact-checked claims from \textit{PolitiFact} and
\textit{Channel4} have been collected
\cite{vlachos-riedel-2014-fact,wang-2017-liar}.

The COVID-19 pandemic has demonstrated the increasing need for
fact-checking in the medical domain. The \textit{FakeCovid} dataset
\cite{shahi-2020-fakecovid} consists of 5,182 fact-checked news
articles mentioning COVID-19, collected from various fact-checking
websites. This dataset however lacks substantiating evidence for
claims. \newcite{kouzy-2020-coronagoesviral} collect 673 COVID-19
related tweets of which 24.8~\% contained misinformation.
\textit{COVID-Fact} \cite{saakyan-et-al-2021} and \textit{HealthVer}
\cite{sarrouti-et-al-2021} both address the limitation of existing
corpora consisting of synthetic or manually summarized claims. Both
datasets provide user-generated Covid-19 related claims.
\textit{COVID-Fact} focuses on Reddit while \textit{HealthVer} claims
stem from excerpts that a search engine produces when queried with
questions about Covid-19.

While most resources rely on existing fact-checks or expert
annotators to label claims, few have explored crowdsourcing to
alleviate the data bottleneck. \newcite{saakyan-et-al-2021} present
crowdworkers with a claim and 5 automatically collected sentences,
from which supporting evidence should be selected, thereby
adjudicating a fact-checking verdict. Further reducing the work of a
crowdworker, \newcite{hanselowski-etal-2019-richly}
task crowdworkers
with refining previously fact-checked claims from \textit{Snopes} by
marking
a specific snippet as evidence within a given source text.  Recently,
\newcite{allen2021crowds} successfully employed crowds in providing
truthfulness ratings for headline and lede sentence pairs from
suspicious articles on Facebook. They find that aggregated crowd
judgments strongly correlate with judgments made by professional
fact-checkers \cite{allen2021crowds}.

To the best of our knowledge there is no existing resource that
facilitates fact-checking user-generated biomedical claims related to
COVID-19 on Twitter. To explore this task, we focus specifically on
creating a corpus and leverage the crowd in making veracity
assessments and retrieving substantiating evidence
to support their verdict.

\section{Corpus Creation and Annotation}

\subsection{Data Collection and Preprocessing}

We sample tweets from an in-house tweet repository based on medical
terms from
MeSH\footnote{\url{https://www.nlm.nih.gov/mesh/meshhome.html}} and
frequently occurring terms. We select data posted between January 2020
and June 2021 containing a mention of COVID-19 and one of the terms
`effect', `side-effect',
`vaccine', `symptom' or `treatment'. Further, we select only tweets
that also contain the lexeme `caus' (matching `causes', `caused',
etc.) to narrow the
search space to `causal' relations.\footnote{We use a boolean
expression with inexact matching: (COVID OR
corona) AND (effect OR side-effect OR vaccine OR symptom OR
treatment) AND (caus*)} This results in a total sample of 38,251
COVID-19-related tweets.

As this set still contains many non-biomedical tweets (e.g.\ tweets
discussing politics around COVID-19), we select a random starting set
of 1118 tweets and manually label them as either `biomedical' or
`non-biomedical', resulting in 408 and 710 per class respectively. We
train a bag-of-words-based feed-forward neural
network\footnote{Implementation details can be found in
the Appendix, Table \ref{tab:nn-filter-params}.} to predict this
binary class for the remaining
tweets. We further filter tweets unlikely to contain a claim using the
claim detection model for tweets from \newcite{wuehrl2021claim}. After
removing duplicates and retweets, we are left with a set of 3,785
biomedical tweets containing claims with causal relations, from which
we randomly sample 300 tweets for our annotation tasks.

\subsection{Annotation}

\subsubsection{Entity and Relation Annotation}

We annotate entities and relations in the biomedical claim tweets for
the downstream verification task. We perform manual annotations
supported by the scispaCy model for
biomedical NER\footnote{See en\_core\_sci\_lg from \url{https://github.com/allenai/scispacy}.}
\cite{neumann-etal-2019-scispacy}. This model assigns a
generic `Entity' label which we manually categorize as one of the
following types:
\begin{compactdesc}
\small
\item[Medical Condition:] Mentions of diseases, illnesses, ailments or disorders.
\item[Treatment:] Medical care given to patients for a disease or illness.
\item[Symptom/Side-effect:] Secondary physical effects of a
medical treatment or condition.
\item[Other:] Relevant entities that do not fall into the above
mentioned categories.
\end{compactdesc}

\begin{figure}
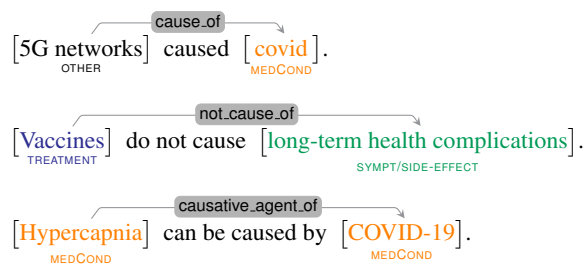

\footnotesize
  \begin{dependency}[edge style={Gray}, label style={gray!60, fill=gray!60,font=\sf,text=black}]
    \begin{deptext}
      \other{5G networks} \& caused \& \medCond{covid}.\\
    \end{deptext}
    \depedge[edge unit distance=0.5ex]{1}{3}{cause\_of}
  \end{dependency}
  
  \hspace{0.3em}
  
  \begin{dependency}[edge style={Gray}, label style={gray!60, fill=gray!60,font=\sf,text=black}]
    \begin{deptext}
      \treatment{Vaccines} \& do not cause \& \sympt{long-term health complications}.\\
    \end{deptext}
    \depedge[edge unit distance=0.5ex]{1}{3}{not\_cause\_of}
  \end{dependency}
  
  \hspace{0.3em}
  
  \begin{dependency}[edge style={Gray}, label style={gray!60, fill=gray!60,font=\sf,text=black}]
    \begin{deptext}
      \medCond{Hypercapnia} \& can be caused by \& \medCond{COVID-19}.\\
    \end{deptext}
    \depedge[edge unit distance=0.5ex]{1}{3}{causative\_agent\_of}
  \end{dependency}
  \caption{Excerpts of tweets that contain named entity and relation annotations.}
  \label{fig:ent_rel_diagram}
\end{figure}

We discard non-biomedical entities that the model incorrectly
identified and only keep entities that are directly relevant to the
causal phrase in the tweet claim. Figure \ref{fig:ent_rel_diagram}
shows examples of annotated tweet claims. To test the quality of this
manual revision, a second annotator labels a random subset of 100
tweets. Both annotators are female, with a background in
(computational) linguistics, aged 27 and 28 years old,
respectively.\footnote{We provide the annotation guidelines and
example fact-checking annotation environment together with the corpus.}

The same annotator who assigned the entity labels further annotates
each instance with relations. We use the relations \textit{cause\_of}
(UR214) and \textit{causative\_agent\_of} (UR173) from UMLS, a medical
terminology, classification and coding system
\cite{umls}, as well as the relation \textit{not\_cause\_of} (UR214!).
The entity and relation annotations are relevant for creating prompts for
language models.

\subsubsection{Fact-checking Annotation}
\paragraph{Annotation Environment.}
We use \textit{Google
  Forms}\footnote{\url{https://www.google.com/forms/about/}} to
collect fact-checking verdicts and substantiating evidence for the
tweet claims and recruit annotators via
\textit{Prolific}\footnote{\url{https://prolific.co/}}. We present 10
tweets with instructions as one annotation task (An example annotation
environment for one tweet can be found in the Appendix). For each
tweet, we ask annotators to search for the claim's key terms using
Google Search to find supporting or refuting sources. Sources should
be credible and reputable (extensions `.gov' and `.mil' and general
news, medical and scientific articles), and avoid resources that are
known to be satirical/comedic like \textit{The
  Onion}\footnote{\url{https://www.theonion.com/}}. If no resource
could be found after searching for two minutes, annotators labeled the
claim as non-verifiable.

If relevant evidence is found, annotators state whether this evidence
\textit{supports} or \textit{refutes} the claim. This leads to three
labels for claim-evidence pairs, answering the question ``Could you
find a resource that confirms of refutes the claim?":
\begin{compactdesc}
\small
\item[\textsc{supports}.] Yes, the resource \textsc{confirms} this claim.
\item[\textsc{refutes}.] Yes, the resource \textsc{refutes} this claim.
\item[\textsc{nei}.] No, I could not find any
  resource that confirms or refutes this claim.
\end{compactdesc}
Annotators are tasked to provide the URL together with the relevant
evidence text snippet.

To ensure that participants follow
instructions carefully, each annotation task consisting of 10 tweets
includes an attention check and reject submissions that
fail this check.

\paragraph{Crowdsourcing Platform and Payment.}
We use Prolific to recruit
annotators. We filter them to meet the following criteria: currently
living in either the UK, US, Ireland or Germany, between 18 and 45
years old, English first language speaker, have achieved at least an
undergraduate degree (BA/BSc/other) in a subject related to
Biomedicine, Biochemistry, Biology or Medicine, and do not have any
literacy difficulties. Each tweet is annotated by three
annotators. The recommended time of completion is 20 minutes.

Each annotator is paid \textsterling7.50 per hour, which corresponds
to \textsterling0.25 for each tweet. To increase awareness of our
task, we implement a bonus system. We manually assess all submission
and award annotators with a bonus of \textsterling0.50 if the
annotations are of good quality, i.e., at least 8 out of 10
annotations are coherent and the given evidence substantiates the
fact-checking label. The expenses of this study amount to
\textsterling314.52.

\section{Corpus Analysis}

\subsection{Annotator Statistics}
\label{ssec:ner_analysis}

78 annotators participated in the fact-checking
annotation. All of the crowdsourced annotators were students at the
time of participation. Even though this was not a strict requirement,
78.2\% have completed an undergraduate and 19.2\% a graduate
degree. Most annotators (96.1\%) are between the age of 19 and 30, and
87.2\% of annotators identified as female, 12.8\% as male. The average
time taken to complete the annotation
task (a set of 10 tweets) was 25:38 minutes, which is slightly longer
than was intended by the instructions.
We presume that participants actually spent 2 minutes verifying each
tweet, and needed additional time to read and understand
instructions.

\subsection{Annotator Agreement and Adjudication}
\subsubsection{Entity and Relation Annotation}

Table~\ref{tab:ia_tweets} presents the inter-annotator (IA) \F scores
achieved between the two annotators manually revising the automatic
named entity annotations. We consider one annotation as gold
annotations and treat the other annotator's labels as predictions
\cite{hripcsak-rothschild_2005}. The agreement between the two
annotators is high, with inter-annotator \F scores above 0.9 in all
but one category (Symptom/Side-effect). The macro-average (weighting
by number of samples per category) is 0.88 \F. Note that the
annotation task did not include segmentation.

\begin{table}
  \centering
  \small
  \begin{tabular}{lc}
    \toprule
    Entity Category & IA \F Score \\
    \cmidrule(lr){1-1} \cmidrule(lr){2-2}
    Treatment & 0.95 \\
    Symptom/Side-effect & 0.68 \\
    Medical Condition & 0.94 \\
    Other & 0.93 \\
    \cmidrule(lr){1-1} \cmidrule(lr){2-2}
    Weighted Macro Average & 0.88 \\
    \bottomrule
  \end{tabular}
  \caption{Inter-annotator (IA) agreement \F scores on the named entities of
    100 tweets.}
  \label{tab:ia_tweets}
\end{table}

\subsubsection{Crowdsourced Fact-checking}

\paragraph{Verification Task.}

We calculate Cohen's $\kappa$ per annotation task (set of 10 tweets),
where each task has three annotators (thus, three pair-wise
comparisons) and take the average of all pairs over all tasks (a total
of 30) as a final agreement score. The average Cohen's $\kappa$ on the
verification labels is 0.44 indicating moderate agreement. However, to
adjudicate the final label for each tweet, we select the majority
decision (where at least 2 annotators agree) as the gold label. Out of
all tweets, only in two cases did annotators not agree at all on the
verification labels. Here, we assign the label \textsc{not enough
  info}. For the remaining tweets, at least two annotators agree on
the verification label.

Most disagreements (55) are between the pair of
labels \textsc{supports} and \textsc{not enough info}. There are only
42 disagreements between the pair \textsc{supports} and
\textsc{refutes}. The pair of labels with fewest disagreements are
\textsc{refutes} and \textsc{nei} (29). This means two annotators are
most likely to disagree on an instance where one annotator labels the
tweet as \textsc{supports} while another annotates it as \textsc{nei},
while disagreements on \textsc{refutes} and \textsc{nei} are not as
common.

\paragraph{Evidence Retrieval.}

To gauge how well the crowdworkers agreed on the evidence they
used to substantiate their verdicts, we compare the URLs they provided
during the annotation process. Out of all 300 tweets, for 78 tweets
(26\% of total) two or more annotators responded with a link to the
same resource.

Notably, even when a link to the same resource was provided by two
different annotators, we observe 5 cases where annotators disagree on
the interpretation of the resource in supporting or refuting the
claim. For instance, the claim ``Covid-19 can cause hearing
impairment, tinnitus and dizziness'' for which annotators provided the
same evidence URL (from
\url{https://www.healthyhearing.com/}) leads to disagreement regarding
\textsc{supports} vs.\ \textsc{refutes} verdicts.

\subsection{Corpus Statistics}

\subsubsection{Entities and Relations}

The resulting CoVERT corpus has a total of 722 entity annotations,
with the number of instances per entity and relation class enumerated
in Table \ref{tab:instances_entity_label}. The largest observed entity
class is `Medical Condition', while the smallest is
`Symptom/Side-effect'. As for the relations, we most frequently
observe the \textit{cause\_of} relation.

\begin{table}[t]
        \centering \footnotesize
        \begin{tabular}{lc}\toprule
            Entity / Relation Class & \# Instances \\
            \cmidrule(lr){1-1} \cmidrule(lr){2-2}
            Medical Condition & 424 \\
            Treatment & 102 \\
	    Other & 102 \\
            Symptom/Side-effect & 94 \\\cmidrule(lr){1-1}\cmidrule(lr){2-2}
	    cause\_of & 253 \\
	    not\_cause\_of & 30 \\
	    causative\_agent\_of & 17 \\
            \bottomrule
        \end{tabular}
        \caption{Instances per entity and relation class in the CoVERT dataset.}
        \label{tab:instances_entity_label}
\end{table}
   
\subsubsection{Crowdsourced Fact-checking}
\paragraph{Verification Task.}

The distribution of labels in the corpus after adjudication is as follows:
198 instances of \textsc{supports}, 66 instances of \textsc{refutes}
and 36 instances of \textsc{nei}.
The \textsc{supports} class has the largest
number of examples, while the \textsc{nei} class has the
least.

\paragraph{Evidence Retrieval.}

\pgfplotstableread[row sep=\\,col sep=&]{
  domain & X-Position & count \\
  cdc & 1 & 104 \\
  nih & 2 &  91 \\
  nature & 3  & 26 \\
  hopkinsmedicine & 4  & 22 \\
  mayoclinic & 5  & 21 \\
  webmd & 6 & 18 \\
  reuters & 7 &  12 \\
  utah & 8 & 10 \\
  who & 9& 9 \\
  harvard & 10 & 9 \\
  europa & 11 & 9 \\
  osu & 12 & 8 \\
  fullfact & 13 & 7 \\
  yalemedicine & 14 & 6 \\
  umn & 15 & 6 \\
  sciencemag & 16 &  6 \\
  medicalnewstoday & 17 & 6 \\
  jamanetwork & 18 & 6 \\
  healthline & 19 &  6 \\
  google &20 & 6 \\
}\domainData

\begin{figure}[t!]
  \centering
  \begin{tikzpicture}
    \scriptsize
    \begin{axis}[
      ybar=5pt,
      bar width=.2cm,
      width=.5\textwidth,
      height=.3\textwidth,
      legend style={at={(0.5,1)},
        anchor=north,legend columns=-1},
      xticklabels={cdc,
        nih,
        nature,
        hopkinsmedicine,
        mayoclinic,
        webmd,
        reuters,
        utah,
        who,
        harvard,
        europa,
        osu,
        fullfact,
        yalemedicine,
        umn,
        sciencemag,
        medicalnewstoday,
        jamanetwork,
        healthline,
        google,
        uofmhealth,
        nytimes,
        muhealth,
        health-desk,
        health,
        clevelandclinic,
        cancer,
        bmj,
        biomedcentral,
        weillcornell},
      xticklabel style={rotate=90},
      xtick=data,
      nodes near coords,
      nodes near coords align={vertical},
      ymin=0,ymax=115,
      xlabel={Domain Names},
      ylabel={\# of occurrences},
      ]
      \addplot table[x=X-Position,y=count]{\domainData};
    \end{axis}
  \end{tikzpicture}
  \caption{Top 20 most frequent domain names occurring in URL links to
    evidence resources.}
  \label{fig:domains}
\end{figure}
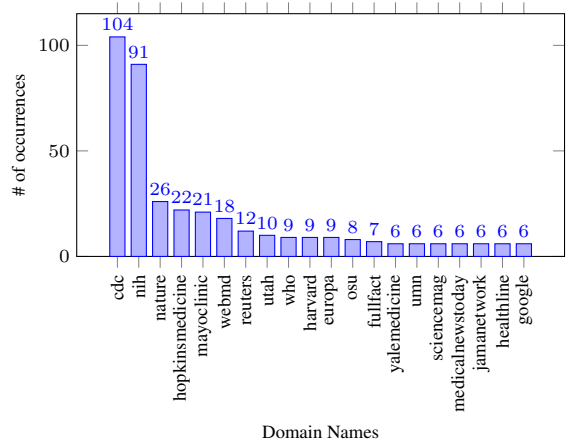

We collect a total of 659 URLs from the annotation task.
Figure \ref{fig:domains} displays the 20
most frequently referenced domain names and their counts, which make
up 66\% of all mentioned URLs. 188 domain names are unique, but
150 of these domains occur less than 5 times in the collected
data.

The most frequently mentioned domain names are medical and health
science related domains. Most of these are also generally deemed
reputable and credible sources of information such as the CDC (Center
for Disease Control and Prevention) and NIH (National Institutes of
Health).  This assures the annotators were following the annotation
instructions carefully, leading to a good quality of annotations.

\section{Experiments}
We want to investigate the extent to which access to external evidence
impacts the prediction of fact-checking verdicts for COVID-19-related
claims from Twitter.  To explore this, we employ a fact-checking
pipeline proposed by \newcite{lee2020language}. Their approach does
not access external evidence. Instead it extracts evidence from the
implicit knowledge contained in the language model BERT
\cite{devlin-etal-2019-bert}. We adapt this system such that it can
access external fact-checking evidence (i.e., evidence annotators
extracted for CoVERT) and compare the verdict prediction performance.

Before investigating this as our primary research objective, we
perform two preliminary experiments. As the claims within the CoVERT
data are Twitter-based instead of Wikipedia-based (FEVER dataset
\cite{Thorne18FEVER}), we first explore to which extend BERT and
BioBERT \cite{lee-2019-biobert} contain domain-specific knowledge
discussed in the CoVERT dataset. For this, we probe each model with
the BioLAMA probe \cite{sung2021language}. In addition, we want to
understand how capable BERT is to ``generate" evidence sentences as
suggested by \newcite{lee2020language} in our setting to be used in
the fact-checking pipeline. We therefore analyze the predictions BERT
makes when unmasking entities in the tweets from our
dataset.  We outline the methodology and results for both preliminary
experiments and the fact-checking pipeline in the following section.

\subsection{Methods}
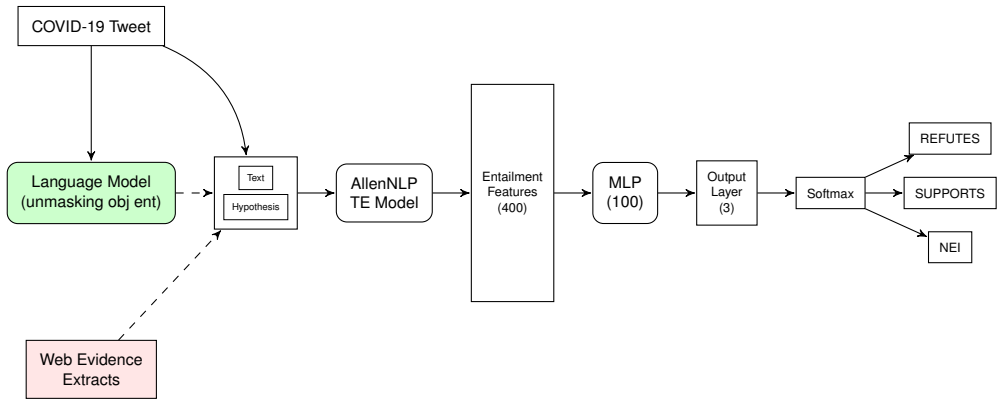
\begin{figure*}
  \centering
  \pdfoutput=1
\makeatletter
\pgfdeclareshape{datastore}{
  \inheritsavedanchors[from=rectangle]
  \inheritanchorborder[from=rectangle]
  \inheritanchor[from=rectangle]{center}
  \inheritanchor[from=rectangle]{base}
  \inheritanchor[from=rectangle]{north}
  \inheritanchor[from=rectangle]{north east}
  \inheritanchor[from=rectangle]{east}
  \inheritanchor[from=rectangle]{south east}
  \inheritanchor[from=rectangle]{south}
  \inheritanchor[from=rectangle]{south west}
  \inheritanchor[from=rectangle]{west}
  \inheritanchor[from=rectangle]{north west}
  \backgroundpath{
    \southwest \pgf@xa=\pgf@x \pgf@ya=\pgf@y
    \northeast \pgf@xb=\pgf@x \pgf@yb=\pgf@y
    \pgfpathmoveto{\pgfpoint{\pgf@xa}{\pgf@ya}}
    \pgfpathlineto{\pgfpoint{\pgf@xb}{\pgf@ya}}
    \pgfpathmoveto{\pgfpoint{\pgf@xa}{\pgf@yb}}
    \pgfpathlineto{\pgfpoint{\pgf@xb}{\pgf@yb}}
 }
}
\makeatother

\begin{tikzpicture}[
  font=\sffamily,
  every matrix/.style={ampersand replacement=\&,column sep=.5cm,row sep=.5cm},
  source/.style={draw,inner sep=.3cm},
  process/.style={draw,diamond,fill=blue!20},
  lm/.style={source,rounded corners,fill=green!20},
  ev/.style={source,fill=pink!40},
  allen/.style={source,rounded corners},
  dots/.style={gray,scale=2},
  to/.style={->,>=stealth',shorten >=0.5pt,font=\sffamily\footnotesize},
  every node/.style={align=center,scale=0.6},
  scale=0.6]

  \matrix{
      \node[source] (tweet) {COVID-19 Tweet}; \\

      \node[lm] (langmodel) {Language Model \\ (unmasking obj ent)};
      \& \node[source, inner sep=.2cm] (pair) {\tikz{
            \node[source] (text) {Text};
            \node[source] (hypothesis) [below=.1cm of text] {Hypothesis};
      }};
      \& \node[allen] (te) {AllenNLP\\ TE Model};
      \& \node[source,scale=0.8,minimum height=6cm] (entail) {Entailment\\Features\\(400)};
      \& \node[allen] (mlp) {MLP\\(100)};
      \& \node[source,scale=0.8] (out) {Output\\Layer\\(3)};
      \& \node[source,scale=0.8] (soft) {Softmax};

      \& \node[source,scale=0.8] (1) {SUPPORTS};
	\node[source,scale=0.8,above = .3cm of 1] (2) {REFUTES};
	\node[source,scale=0.8,below = .3cm of 1] (0) {NEI}; \\

      \node[ev] (evidence) {Web Evidence\\Extracts};\\
  };

  \draw[to] (tweet) -- node[midway,below] {}
      node[midway,above] {} (langmodel);
  \draw[to,dashed] (langmodel) -- node[midway,below] {}
      node[midway,left] {} (pair);
  \draw[to,dashed] (evidence) -- node[midway,left] {} (pair);
  \draw[to] (tweet) to[bend left=30] node[midway,above] {}
      node[midway,below] {} (pair);
  \draw[to] (pair) -- node[midway,right] {}
      node[midway,left] {} (te);
  \draw[to] (te) -- node[midway,right] {}
      node[midway,left] {} (entail);
  \draw[to] (entail) -- node[midway,right] {}
      node[midway,left] {} (mlp);
  \draw[to] (mlp) -- node[midway,right] {}
      node[midway,left] {} (out);
  \draw[to] (out) -- node[midway,right] {}
      node[midway,left] {} (soft);
  \draw[to] (soft) -- node[midway,right] {}
      node[midway,left] {} (0);
  \draw[to] (soft) -- node[midway,right] {}
      node[midway,left] {} (1);
  \draw[to] (soft) -- node[midway,right] {}
      node[midway,left] {} (2);
\end{tikzpicture}
  \caption{Full depiction of the fact-checking
    pipeline. `Hypothesis' is either generated by the language model (green)
    or taken from crowdsourced annotations (red).}
  \label{fig:model_diagram}
\end{figure*}
\subsubsection{Probing Language Models for Domain-specific
Knowledge}
\label{probe_LM_domain}
To gauge how well implicit knowledge is stored in BERT and BioBERT,
and how well these language models lend themselves as a source of
evidence like \newcite{lee2020language} suggest, we investigate to
which extent information from the CoVERT dataset is contained within
these two language models. We employ the BioLAMA probe suggested by
\newcite{sung2021language} which allows us to see whether a language
model is able to correctly predict masked object entities in a
constrained setting. BioLAMA originally probes BERT and BioBERT for
their inherent relational knowledge using factual triples sourced from
CTD, UMLS and Wikidata. Comparing probing results for CoVERT data to
those attained using the original dataset of biomedical factual
triples in \newcite{sung2021language} gives insights into the domain
of biomedical tweets and allows us to conclude which language model
best captures the domain-specific knowledge within CoVERT.

BioLAMA generates fill-in-the-blank cloze statements or `prompts' like
``\underline{Hepatitis} has symptoms such as [Y].'', where [Y] is the
masked object, unmasked as `abdominal pain' by BioBERT.  In our
experiment, we use the manual approach for generating prompts as
described in \newcite{sung2021language}.  We follow their evaluation
methodology and use top-\textit{k} accuracy as the evaluation
metric. This is equal to 1 for an instance if any of the
top-\textit{k} object entities match an object in the gold annotated
object list. If there are no matches, the score for this instance is
0. This binary setting allows to calculate accuracy as the number of
correct predictions devided by all predictions.

\subsubsection{Language Model Capacity for Unmasking Claim Entities}

In addition to probing the knowledge within the language model, we
further want to understand its capacity to create `evidence texts' for
modelling the fact-checking task without access to external evidence.
Therefore, we conduct an analysis of the predictions BERT makes when
masking object entities in the biomedical claim tweets. For each of
the three fact-checking categories, we extract and analyze the
probabilities with which the object entity is predicted. We report the
probability as 0 if the correct entity does not appear in the top 1000
predictions.

\subsubsection{Verdict Prediction With and Without External Evidence}

We investigate whether verdict prediction benefits from access to
external evidence, or whether this evidence can be replaced with
unmasked claims from a language model and still achieve similar
results. To do this, we re-implement the fact-checking pipeline by
\newcite{lee2020language}. Figure \ref{fig:model_diagram} shows a
diagram of the pipeline. The input to the pipeline consists of a
$\langle$\textit{text, hypothesis}$\rangle$ pair, which is passed to
the textual entailment (TE) model from
AllenNLP \cite{gardner-2018-allennlp}. We take the last layer of the
pretrained entailment model (before the softmax) to obtain
``entailment features'', which are passed on to a multi-layer
perceptron (MLP) for the final verification prediction. The MLP
component is originally trained on the FEVER 2018 training set
\cite{Thorne18FEVER}, and is referred to as MLP-FEVER in our
experiments.

For the \textit{hypothesis} component of the input pair, we experiment
with two types of inputs. First, we use the approach by
\newcite{lee2020language} and generate the \textit{hypothesis} using
the language model BERT \cite{devlin-etal-2019-bert}, which unmasks an
entity in the original text. Alternatively, we input the respective
evidence snippet from the CoVERT dataset, effectively giving the
pipeline access to external, real-world evidence. Refer to Table
\ref{tab:te_instance} for an example of such a $\langle$\textit{text,
hypothesis}$\rangle$ pair.

To further test the effect of evidence extracts on the pipeline, we
additionally fine-tune the MLP component with text and evidence pairs,
which we call MLP-Evidence in our experiments, using a 80/10/10
train--develop--test split of the CoVERT data.\footnote{Implementation
details and hyperparameters of MLP-FEVER and MLP-Evidence can be found
in the Appendix, Table \ref{tab:mlp-params}.}

\begin{table}
  \centering
  \small
  \begin{tabularx}{\columnwidth}{clX}
    \toprule
    &Text & ``Covid causes \underline{heart inflammation} in like 70\% of people (including asymptomatic).  That seems like a high rate of heart inflammation.'' \\
    \cmidrule(lr){1-2}\cmidrule(lr){3-3}
    \multirow{10}{*}{\rotatebox{90}{Hypotheses}} & LM & ``Covid causes
    \underline{inflammation} in like 70\% of people ( including asymptomatic ). That seems like a high rate of heart inflammation.'' \\
    \cmidrule(lr){2-2} \cmidrule(lr){3-3}
    &Web & ``The SARS-CoV-2 virus can damage the heart in several ways. For example, the virus may directly invade or inflame the heart muscle, and it may indirectly harm the heart by disrupting the balance between oxygen supply and demand.'' \\
    \bottomrule
  \end{tabularx}
  \caption{Example instance for verdict prediction. The
    \textit{text} is the original tweet, while the
    \textit{hypothesis} is either the tweet unmasked by the
    language model (LM) or the evidence extract retrieved from the
    web. The underlined entity is unmasked by the language model.}
  \label{tab:te_instance}
\end{table}

\subsection{Results}

\subsubsection{BioLAMA Probing}

\begin{table}
  \centering\small
  \begin{tabular}{lrrrr}
    \toprule
    & \multicolumn{2}{c}{BERT} & \multicolumn{2}{c}{BioBERT} \\ 
    \cmidrule(lr){2-3}\cmidrule(lr){4-5}
    Dataset & Acc@1 & Acc@5 & Acc@1 & Acc@5 \\ \midrule
    \textsc{supports} & 4.6\phantom{0} & 10.79 & 1.59 & 5.56\\
    \textsc{refutes} & 3.62 & 6.57 & 2.17 & 5.0\phantom{0} \\
    \textsc{nei} & 0.0\phantom{0} & 7.07 & 0.0\phantom{0} & 25.18 \\\midrule
    \textsc{ctd+umls} & & & & \\
    +Wiki & 0.86 & 3.08 & 1.75 & 6.09 \\\bottomrule
  \end{tabular}
  \caption{Accuracy scores (\%) for BERT probed with
    BioLAMA using CoVERT (\textsc{supports, refutes} and
    \textsc{nei}) and the original BioLAMA probing
    collection (CTD, UMLS and Wikidata).}
  \label{tab:biolama-results}
\end{table}

We employ the BioLAMA probe to investigate whether BERT and BioBERT
contain suitable domain-specific knowledge to serve as an evidence
source during fact-checking. Table \ref{tab:biolama-results} reports
results for the probe within the CoVERT dataset as well as the scores
for the original BioLAMA probe \cite{sung2021language}. The
\textsc{supports} class is most accurately modeled out of the three
verdict classes (4.6 Acc@1), while the lowest score (0.0 Acc@1) is
achieved for \textsc{nei}.

Generally, BERT achieves higher scores on the CoVERT data than the
original corpus of triples.  The original BioLAMA probe using BioBERT
on the CTD, UMLS and Wikidata collection achieved a score of 1.75 Acc@1
and 6.09 Acc@5. On the CoVERT corpus, BERT achieves higher scores than
BioBERT, even though the corpus consists of biomedical information. We
hypothesize that this is because the specialized biomedical language
used in the BioBERT training data might not be a good match for the
non-expert language commonly used in tweets. Since
BERT appears better suited for our dataset, we use this language model
in subsequent experiments.

\subsubsection{Language Model Capacity for Unmasking Claim Entities}
We further explore how capable the BERT language model is to unmask
entities in biomedical claims. Table \ref{tab:five_point_analysis}
reports the probabilities with which masked object entities are
predicted by BERT. We report the results grouped by the fact-checking
class (\textsc{supports}, \textsc{refutes} and \textsc{nei}) that each
instance belongs to.

Most entities are predicted with very low probability
by BERT. There are differences between the three
fact-checking categories, where \textsc{supports} has a third quartile
that is higher than that of \textsc{refutes}. Additionally, the
\textsc{nei} category has a maximum lower than that of
\textsc{supports} and \textsc{refutes}. However, the mean, median and
first quartile of all three categories are all very close or equal to
0. The language model still predicts the object entity of incorrect
claims with high probabilities. This indicates that predictions for
this class have to be considered carefully when using it in the
downstream task of evidence creation.

\begin{table}
  \centering\small
  \begin{tabular}{llllll}\toprule
    Class & Min & Q1 & Q2 &
                            Q3 & Max\\
    \cmidrule(lr){1-1}\cmidrule(lr){2-2}\cmidrule(lr){3-3}\cmidrule(lr){4-4}\cmidrule(lr){5-5}\cmidrule(lr){6-6}
    \textsc{sup.} & 0 & 0 & 0.00057 & 0.03258 & 0.98289 \\
    \textsc{ref.} & 0 & 0 & 0 & 0.00210 & 0.98672 \\
    \textsc{nei} & 0 & 0 & 0.00003 & 0.00995 & 0.7943 \\
    \bottomrule
  \end{tabular}
  \caption{Five-number summary of probabilities with which masked object entities are predicted by BERT.}
  \label{tab:five_point_analysis}
\end{table}

\subsubsection{Verdict Prediction With and Without External Evidence}
Our main research question is to investigate the impact of external
evidence when predicting fact-checking verdicts for COVID-19-related
claims in tweets.
Table~\ref{tab:results-models} presents the results from the
verdict prediction experiment. We report the results for
two prediction pipelines, namely MLP-FEVER (following
\newcite{lee2020language}) and our adaptation of this pipeline
(MLP-Evidence) in which the MLP component is fine-tuned with evidence
and text pairs from CoVERT, giving it access to external
evidence. Each pipeline is evaluated on the FEVER 2018 test dataset
\cite{Thorne18FEVER} consisting of 9999 instances, and on pairs of
tweets and language model generated evidence (Tweet + LM) and tweets
and CoVERT evidence (Tweet + Evidence).

Provided with hypotheses generated using BERT, MLP-FEVER achieves the
highest \F score of 0.60.  The same pipeline achieves an 0.49 \F
on the FEVER dataset. When using the evidence extracts as hypotheses,
the performance drops to 0.46 \F.
Fine-tuning the MLP component with evidence extracts (MLP-Evidence)
the performance is slightly lower on FEVER data and the Tweet + LM
pairs (difference of .03 \F, respectively), however, achieves
substantially higher \F score of 0.69 on the CoVERT corpus.

The results show that fine-tuning the MLP component with
evidence extracts, the pipeline achieves higher scores
than with access to `evidence' generated by BERT only, showing
that the retrieved evidence extracts contribute to the performance of
this pipeline.

\begin{table}
  \centering\small
  \setlength{\tabcolsep}{3pt}
  \begin{tabular}{lccccccccc}
    \toprule
    \multirow{2}{*}{} & \multicolumn{3}{c}{FEVER} & \multicolumn{3}{c}{Tweet + LM} & \multicolumn{3}{c}{Tweet + Ev.}\\
    \cmidrule(lr){2-4}\cmidrule(lr){5-7}\cmidrule(lr){8-10}
                      &P&R&\F &P&R&\F &P&R&\F\\
    \cmidrule(r){2-2}\cmidrule(lr){3-3}\cmidrule(lr){4-4}\cmidrule(lr){5-5}\cmidrule(lr){6-6}\cmidrule(lr){7-7}\cmidrule(lr){8-8}\cmidrule(lr){9-9}\cmidrule(l){10-10}
    MLP-FEVER & .52 &    .47    &  \textbf{.49} &  .60   &   .61  &    \textbf{.60}  &   .61   &   .38   &   .46 \\
    MLP-Evidence & .64  &    .36    &  .46 & .66   &   .68  &    .57 & .68 &  .74    &  \textbf{.69}
    \\
    \bottomrule
  \end{tabular}
  \caption{Results for the two fact-checking pipelines, MLP-FEVER and MLP-Evidence,
    evaluated on the FEVER 2018 dataset, Tweet + LM Pairs and Tweet + Evidence Pairs. The highest \F score is highlighted for each input set.}
  \label{tab:results-models}
\end{table}

\subsection{Error Analysis and Discussion}

    To see how predictions change when the pipeline has access to
    language model generated evidence or CoVERT evidence extracts, we
    conduct a qualitative error analysis into predictions made by the
    fact-checking pipeline before and after fine-tuning.

    We observe that MLP-FEVER with BERT-generated evidence mistakenly
    labels 4 out of 6 instances as \textsc{refutes} that are actually
    \textsc{supports} instances. Inspecting these cases, we find that
    the language model generated evidence text is likely to produce
    features indicating
    \textit{entailment}, as can be seen in Ex.~\textit{b}, Table
    \ref{tab:lm-errors} in the Appendix. Similarly, in 3 out of 15
    \textsc{supports} instances are mistakenly labelled
    \textsc{refutes} by MLP-FEVER, with example \textit{c} in Table
    \ref{tab:lm-errors} showing that the evidence created by BERT is
    sufficiently dissimilar to result in
    \textit{contradiction} entailment features in our pipeline,
    finally resulting in \textsc{rerfutes} predictions. These
    instances show that BERT is creating evidence texts that are
    likely unuseful to the pipeline.

    The pipeline struggles to make use of evidence extracts
    when the MLP component is only trained on the FEVER dataset.
    We inspect 10 instances where MLP-FEVER incorrectly predicts
    \textsc{nei} instead of \textsc{supports} and find that the
    evidence extracts are much longer (and sometimes a long
    paragraph), thus including more information than hypotheses
    originally seen in FEVER. Examples of this can be seen in Table
    \ref{tab:ev-errors}, Ex.~\textit{d} and \textit{f}. In both
    instances, the MLP-Evidence pipeline (fine-tuned with evidence
    from CoVERT) correctly predicts \textsc{supports}, showing that
    the MLP component is able to interpret the input pairs correctly
    after fine-tuning.

    Although the overall \F score of the pipeline increases when
    fine-tuned with evidence extracts from CoVERT, the total number of
    \textsc{supports} class predictions has increased from 33 to 55,
    with 15 false positives, and
    very seldomly predicting the \textsc{nei} class. This may be an
    inherent limitation of the evidence extracts, as they by
    definition do not contain any evidence texts for the \textsc{nei}
    class, meaning this class is never modelled during fine-tuning.
    Additionally, the distribution across classes in the fine-tuning
    set is unbalanced (422 evidence extracts for \textsc{supports} vs.
    128 for \textsc{refutes}).

\section{Conclusion and Future Work}
    We present CoVERT, the first fact-checked biomedical
    COVID-19-related tweet corpus, along with a novel approach to
    using evidence extracts as part of a verdict prediction pipeline.
    We outline how we leverage a crowd for fact-checking, finding
    moderate agreement among annotators and reliable annotations when
    aggregating verdict labels. Our extension of the verdict
    prediction pipeline \cite{lee2020language} using evidence extracts
    from CoVERT indicates that the task can benefit from real-world
    evidence rather than only using `evidence' generated from the
    implicit knowledge contained in language models.

    Apart from facilitating evidence-based fact-checking directly, the
    CoVERT corpus additionally allows querying structured databases
    for evidence retrieval and verdict prediction, as it offers
    annotated entities and relations, which can be linked to
    ontologies.

   Future work may consider if the pipeline benefits from more
    fine-grained evidence extracts. Evidence extracted at sentence
    level \cite{hanselowski-etal-2019-richly} may contain information
    directly relevant to the claim, thereby facilitating richer
    entailment features.
    
\section{Ethical Considerations}
All participants in the crowdsourcing study agreed to participate at
their own volition and signed a consent form at the outset of the
study. Although we did not consider the task to present any harm to
the crowd workers, it is possible that annotators were exposed to
false information in the tweets and during their research.

It is important that the annotations in this dataset are not
taken out of context with regard to the time-frame at which they were
annotated.  As biomedical knowledge, particularly with regard to
SARS-COV-2, is continuously updated as new research is published, the
evidence extracts and verdicts in the CoVERT dataset may be outdated
in the future.

\section{Acknowledgements}
This research has been conducted as part of the FIBISS project which
is funded by the German Research Council (DFG, project number: KL
2869/5-1).
We thank Enrica Troiano for valuable discussions.

\section{Bibliographical References}\label{reference}

\bibliographystyle{lrec2022-bib}
\bibliography{lit}

\onecolumn

\section*{Appendix}

\subsection*{Neural Network for Biomedical Filtering}

\begingroup
\small
\begin{tabular}{ll}
  \toprule
  Parameter/Variable & Setting \\\cmidrule(l){1-1} \cmidrule(lr){2-2}
  Input Size & 5000 (BOW vector) \\
  Hiden Size & 100 \\
  Output Layer Size & 2 \\
  Activation Function & Sigmoid \\
  Optimizer & Adam \\
  Learning Rate & 0.01 \\
  Loss Function & Cross Entropy \\
  Training epochs & 30 \\
  \# Train/Dev/Test Data & 894/111/111 \\
  \bottomrule
\end{tabular}
\captionof{table}{Implementation details and training parameters of the feed-forward neural network for biomedical filtering.}
\label{tab:nn-filter-params}
\endgroup

\vspace{1cm}

\subsection*{MLP Component of the Verdict Prediction Pipeline}

\begingroup
\small
\begin{tabular}{lll}
  \toprule
  & MLP-FEVER & MLP-Evidence \\
  Parameter/Variable & Setting & Setting \\
  \cmidrule(lr){1-1}\cmidrule(lr){2-2}\cmidrule(l){3-3}
  Input Size & 400 & 400 \\
  Hiden Size & 100 & 100 \\
  Output Layer Size & 3 & 3 \\
  Activation Function & ReLU & ReLU \\
  Optimizer & Adam & Adam \\
  Learning Rate & 0.001 & 0.01 \\
  Loss Function & Cross Entropy & Cross Entropy \\
  Max training epochs & 200 & 120 \\
  Patience & 30 & 10 \\
  Batch size & 32 & 32 \\
  \# Train/Dev/Test Data & 116~359/14~544/14~544 & 439/6/6\\
  \bottomrule
\end{tabular}
\captionof{table}{Implementation details and training/fine-tuning parameters of the MLP
  component in the verdict prediction pipeline. MLP-FEVER: training on FEVER 2018 dataset,
  MLP-Evidence: fine-tuning on CoVERT evidence extracts.}
\label{tab:mlp-params}
\endgroup

\vfill
\mbox{}

\pagebreak

\subsection*{Error Analysis: Predictions on Language Model `Evidence'}

\begingroup
\centering\small
\begin{tabularx}{\linewidth}{lXccccc}
  \toprule
  &&&& \multicolumn{3}{c}{Classification} \\
  \cmidrule(lr){5-7}
  && \multicolumn{2}{c}{Entity} & & \multicolumn{2}{c}{Prediction} \\
\cmidrule(lr){3-4}\cmidrule(lr){6-7}
  ID & Tweet & Masked & Pred. & Gold & MLP-FEVER & MLP-Evidence \\
\cmidrule(r){1-1}\cmidrule(lr){2-2}\cmidrule(rl){3-3}\cmidrule(lr){4-4}\cmidrule(lr){5-5}\cmidrule(lr){6-6}\cmidrule(l){7-7}
  \textit{a} & Stop calling it a vaccine!! Vaccines contain the same
               germs that cause \underline{disease}. & disease & measles &                                                                            R & R &
                                                                                                                                                           S \\
\cmidrule(r){1-1}\cmidrule(lr){2-2}\cmidrule(rl){3-3}\cmidrule(lr){4-4}\cmidrule(lr){5-5}\cmidrule(lr){6-6}\cmidrule(l){7-7}
            \textit{b} & Covid-19 vaccines initiates an early and progressive
	    clotting of blood in the lungs (pulmonary thrombosis) which impairs
	    blood supply and gas exchange at lungs, leading to respiratory
	    failure, which in majority of cases cause \underline{death}. & death &
	    death & R &
	    S & S
  \\
\cmidrule(r){1-1}\cmidrule(lr){2-2}\cmidrule(rl){3-3}\cmidrule(lr){4-4}\cmidrule(lr){5-5}\cmidrule(lr){6-6}\cmidrule(l){7-7}
            \textit{c} & If it's unclear it can't be reported.  The death has to
	    be contributed from Covid because Covid causes the
	    \underline{respiratory issues} etc that cause the death, Covid caused
	    it.  Like lung cancer if death because their lung stopped
	    functioning because of cancer they died from lung cancer. &
	    respiratory issues & cancer & S &
	    R & S
	    \\\bottomrule
\end{tabularx}
\captionof{table}{Instances where the pipeline failed
  to correctly classify tweets given evidence generated by BERT. R: \textsc{refutes}, S: \textsc{supports}, N: \textsc{not enough information}, MLP-FEVER: training on FEVER 2018 dataset,
  MLP-Evidence: fine-tuning on CoVERT evidence extracts. Masked entities in the
  tweet are underlined.}
  \label{tab:lm-errors}
\endgroup

\vfill

\subsection*{Predictions on Evidence Extracts}
\begingroup
\newlength{\firstcolumnlength}
\setlength{\firstcolumnlength}{0.27\linewidth}
\centering\small
  \begin{tabularx}{\linewidth}{p{\firstcolumnlength}cXccc}
    \toprule
    &&&& \multicolumn{2}{c}{Prediction} \\
    \cmidrule(l){5-6}
    Tweet & ID & Evidence& Gold & MLP-FEVER & MLP-Evidence \\
    \cmidrule(r){1-1}\cmidrule(lr){2-2}\cmidrule(rl){3-3}\cmidrule(lr){4-4}\cmidrule(lr){5-5}\cmidrule(l){6-6}
    \multirow{2}{*}{\parbox{\firstcolumnlength}{Stop calling it a vaccine!! Vaccines contain the same germs that
    cause disease.}}
    & \textit{a} &
    mRNA vaccines teach our cells how to make a protein, or even
    just a piece of a protein that triggers an immune response
    inside our bodies.
    & R & S & S \\
    & \textit{b} &
    a preparation that is administered (as by injection)
    to stimulate the body's immune response against a specific
    infectious agent or disease: such as a preparation of genetic
    material (such as a strand of synthesized messenger RNA) that is
    used by the cells of the body to produce an antigenic substance
    (such as a fragment of virus spike protein)
    & R & S & S\\
    \cmidrule(r){1-1}\cmidrule(lr){2-2}\cmidrule(rl){3-3}\cmidrule(lr){4-4}\cmidrule(lr){5-5}\cmidrule(l){6-6}
    \multirow{2}{*}{\parbox{\firstcolumnlength}{Covid causes heart inflammation in like 70\% of people (including
    asymptomatic).  That seems like a high rate of heart
    inflammation..}}
    & \textit{c} &
    COVID has been associated with a higher incidence of heart
    inflammation in adolescents and young adults.
    & S & S & S \\
    & \textit{d} & The SARS-CoV-2 virus can damage the heart in several
                   ways. For example, the virus may directly invade or inflame the
                   heart muscle, and it may indirectly harm the heart by disrupting the
                   balance between oxygen supply and demand.
          & S & N & S\\
    \cmidrule(r){1-1}\cmidrule(lr){2-2}\cmidrule(rl){3-3}\cmidrule(lr){4-4}\cmidrule(lr){5-5}\cmidrule(l){6-6}
    \multirow{2}{*}{\parbox{\firstcolumnlength}{If it's unclear it can't be
    reported.  The death has to be contributed from Covid because
    Covid causes the respiratory issues etc that cause the death,
    Covid caused it.  Like lung cancer if death because their lung
    stopped functioning because of cancer they died from lung cancer.}}
    & \textit{e} & Most people infected with the COVID-19 virus will experience mild to moderate respiratory illness & S & S & S \\
    & \textit{f} & COVID-19 is a respiratory disease, one that
                   especially reaches into your respiratory tract, which includes your
                   lungs. COVID-19 can cause a range of breathing problems, from mild
                   to critical. & S & N & S \\[2.2cm]

    \bottomrule
  \end{tabularx}
  \captionof{table}{Instances where the pipeline failed to
    correctly classify Tweet/Evidence pairs. R: \textsc{refutes}, S:
    \textsc{supports}, N: \textsc{not enough information}. MLP-FEVER: pipeline with MLP trained only on FEVER,
    MLP-Evidence: pipeline with MLP fine-tuned on evidence extracts from CoVERT. Source of evidence can be found in the corpus file.}
  \label{tab:ev-errors}
\endgroup

\newpage
\subsection*{Example Fact-Checking Annotation Environment}\label{app:annotation_env}

\vspace{1em}

Instructions:
\vspace{1em}

\noindent\fbox{
    \parbox{\textwidth}{
    Fact-check the emboldened claim by
    \begin{enumerate}
        \item reading the tweet carefully
        \item taking note of the claim in bold text
        \item entering its key terms into Google Search
        \item finding a reputable source that confirms or refutes the claim
        \item mark the claim as being confirmed or refuted in the multiple choice question
        \item enter the URL to the reputable source you have found into the given answer slot
        \item copy and paste the segment of text from the source that explicitly confirms or refutes the claim
    \end{enumerate}
    Each claim should take no more than 2 minutes to completely. The bonus is only awarded if a sensible URL and supporting text are pasted into the relevant fields. If you have investigated the Google Search results and are not able to find a confirmation/disconfirmation, please select the multiple choice option ``No, I could not find any resource that confirms or refutes this claim".
    }
}

\vspace{1em}

Tweet:

\vspace{1em}

\noindent\fbox{
    \parbox{\textwidth}{
    Stop calling it a vaccine! \textbf{Vaccines contain the same germs that cause disease.} (For example, measles vaccines contains measles virus, and Hib vaccine contains Hib bacteria.) But they have been either killed or weakened to the point that they don't make you sick. Covid shot doesnt
    }
}

\vspace{1em}

\noindent\fbox{
    \parbox{\textwidth}{
    Could you find a resource that confirms or refutes the claim in bold text?
    
    \begin{todolist}
        \item Yes, the resource CONFIRMS this claim
        \item Yes, the resource REFUTES this claim
        \item No, I could not find any resource that confirms or refutes this claim
    \end{todolist}
    }
}

\vspace{1em}

\noindent\fbox{
    \parbox{\textwidth}{
    Enter the URL to the reputable resource you have found that substantiates or refutes the claim from the question above. If you marked ``No, I could not find any resource \ldots", you do not need to add a URL.
    
    \vspace{1em}
    \fbox{\hspace{0.9em}\color{gray} Enter URL here \hspace{0.9em}}
 }
}

\vspace{1em}

\noindent\fbox{
    \parbox{\textwidth}{
        Enter a short extract from this source that confirms or refutes the above claim.
        
    \vspace{1em}
    \fbox{\hspace{0.9em}\color{gray} Enter text here \hspace{0.9em}}
 }
}

\end{document}